\documentclass[runningheads,a4paper]{llncs}
\usepackage{amsmath}
\usepackage{amssymb}
\usepackage{subfigure}
\usepackage{graphicx}
\usepackage{multirow}
\usepackage{multicol}
\usepackage{color}

\usepackage{url}

\urldef{\mailsa}\path|{a.b, c.d, e.f, g.h|
\urldef{\mailsb}\path|i.j, k.l, m.n, o.p,|
\urldef{\mailsc}\path|emails}@urv.cat|

\begin{document}
\mainmatter  

\title{REFUGE CHALLENGE 2018-Task 2:\\ Deep Optic Disc and Cup Segmentation in Fundus Images Using U-Net and Multi-scale Feature Matching Networks}

\titlerunning{Lecture Notes in Computer Science: Authors' Instructions}

%
%

\author{Vivek Kumar Singh\inst{1,\thanks{Corresponding Author: vivekkumar.singh@urv.cat}} \and Hatem A. Rashwan\inst{1}  \and Adel Saleh\inst{1} \and Farhan Akram\inst{2}  \and Md. Mostafa Kamal Sarker\inst{1} \and Nidhi Pandey\inst{1} \and Saddam Abdulwahab\inst{1}  }

\authorrunning{V.K. Singh et al.}
%
\institute{DEIM, Universitat Rovira i Virgili, Spain. \and
Imaging Informatics Division, Bioinformatics Institute, Singapore.
}




%
%

\maketitle
\begin{abstract}
In this paper, an optic disc and cup segmentation method is proposed using U-Net followed by a multi-scale feature matching network. The proposed method targets task 2 of the REFUGE challenge 2018. In order to solve the segmentation problem of task 2, we firstly crop the input image using single shot multibox detector (SSD). The cropped image is then passed to an encoder-decoder network with skip connections also known as generator. Afterwards, both the ground truth and generated images are fed to a convolution neural network (CNN) to extract their multi-level features. A dice loss function is then used to match the features of the two images by minimizing the error at each layer. The aggregation of error from each layer is back-propagated through the generator network to enforce it to generate a segmented image closer to the ground truth. The CNN network improves the performance of the generator network without increasing the complexity of the model.
\end{abstract}

\section{Introduction}

Retinal fundus image analysis is crucial for ophthalmologist to deal with the medical diagnosis, screening and treatment of opthalmologic diseases. The morphology of the optic disk (OD) and optic cup (OC) is an important structural indicator for assessing the presence and severity of retinal diseases, such as diabetic retinopathy, hypertension, glaucoma, hemorrhages, vein occlusion, and neovascularization \cite{macgillivray2014retinal}. The OD shape and colors are similar to hard exudates which is one of the main signs of diabetic retinopathy. Being able to detect the OD significantly decreases the difficulties to detect the hard exhaudes in the fundus image \cite{saleh2018learning}.
The OD and OC segmentation is the first step for a significant investigation of retinal images that helps in treating eye diseases \cite{almazroa2015optic}. 

\begin{figure}[htp]
\centering
\includegraphics[width=0.35\textwidth, height=0.2\textheight]{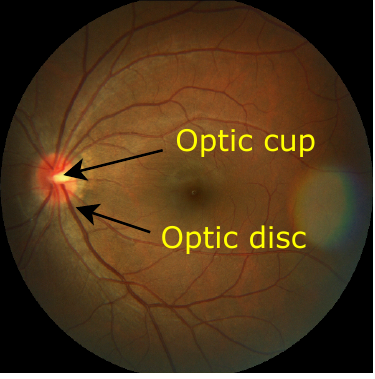}
\caption{Relevant structures in a fundus image.}
\label{fig:figD}
\end{figure}

In color fundus images, OC appears as a bright yellowish oval region, in turn, OD is darker. Fig~\ref{fig:figD} shows an example of a color retinal fundus image with the key anatomical structures denoted. For ophthalmologists and eye care specialists, an automated segmentation and analysis of fundus optic disc plays an important role to diagnose and treat the retinal diseases. 


Numerous methods has been proposed to detect and segment OD and OC. Segmenting of the OC region from fundus images is a challenge due to the low contrast boundary.
In \cite{wong2008level}, an automatic OC segmentation method based on a variational level set was proposed. For diagnosis of glaucoma disease, Chrastek et al.\cite{chrastek2005automated} proposed an automated segmentation algorithm to segment the optic nerve head. They firstly removed the blood vessel by using a distance map algorithm and a morphological operation, and then anchored active contour model has been used to segment the OC region.
With the widespread of using deep learning models in segmentation tasks, many methods have recently been proposed based on convolutional neural network (CNN). An automatic OC and OD segmentation has been proposed in \cite{al2018multiscale} based on a stack of deep U-Net models. Each model in a cascade refines the result of the previous one. In addition, \cite{fu2018joint} proposed a multi-scale deep model with multi-level loss for segmenting OD and OC regions in fundus images. 

In this paper, we propose a retinal joint OD and cup segmentation model based on a U-Net network including encoder and decoder network followed by a CNN network to matching the features of the predicted and ground-truth images to get a segmented image more close to the correct one. The second CNN network is conditioned by the color input image for learning the statistical invariant features (texture, color etc.), in addition to the shape information of the segmented image. Indeed, the second CNN encourages the generator to produce output that cannot be distinguished from ground-truth ones.

The rest of the paper is organized as follows. Firstly, section 2 describes the methodology of the proposed model. In addition, section 3 shows the experiments and results. Finally, the conclusion are explained in section 4.

\section{Material and Methods}

\subsection{REFUGE 2018: Dataset description}
All retinal fundus images were download from REFUGE 2018 challenge \footnote{https://refuge.grand-challenge.org/home/}. We participated the Task 2: Optic disc and cup Segmentation. The dataset is divided into two sets: training (400 images) and validation (400 images). The 400 images of the training set  originally are in JPG format. The training set also includes corresponding 400 ground truth segmentations in BMP format. All images have size of $2124\times 2056$. Validation set is used for on-line self-evaluation and for the final on-site challenge to result the performance rank of different research groups.

\label{subsec:massSegcGAN}
\begin{figure*}[htp]
\centering
\includegraphics[width=1.0\textwidth, height=0.4\textheight]{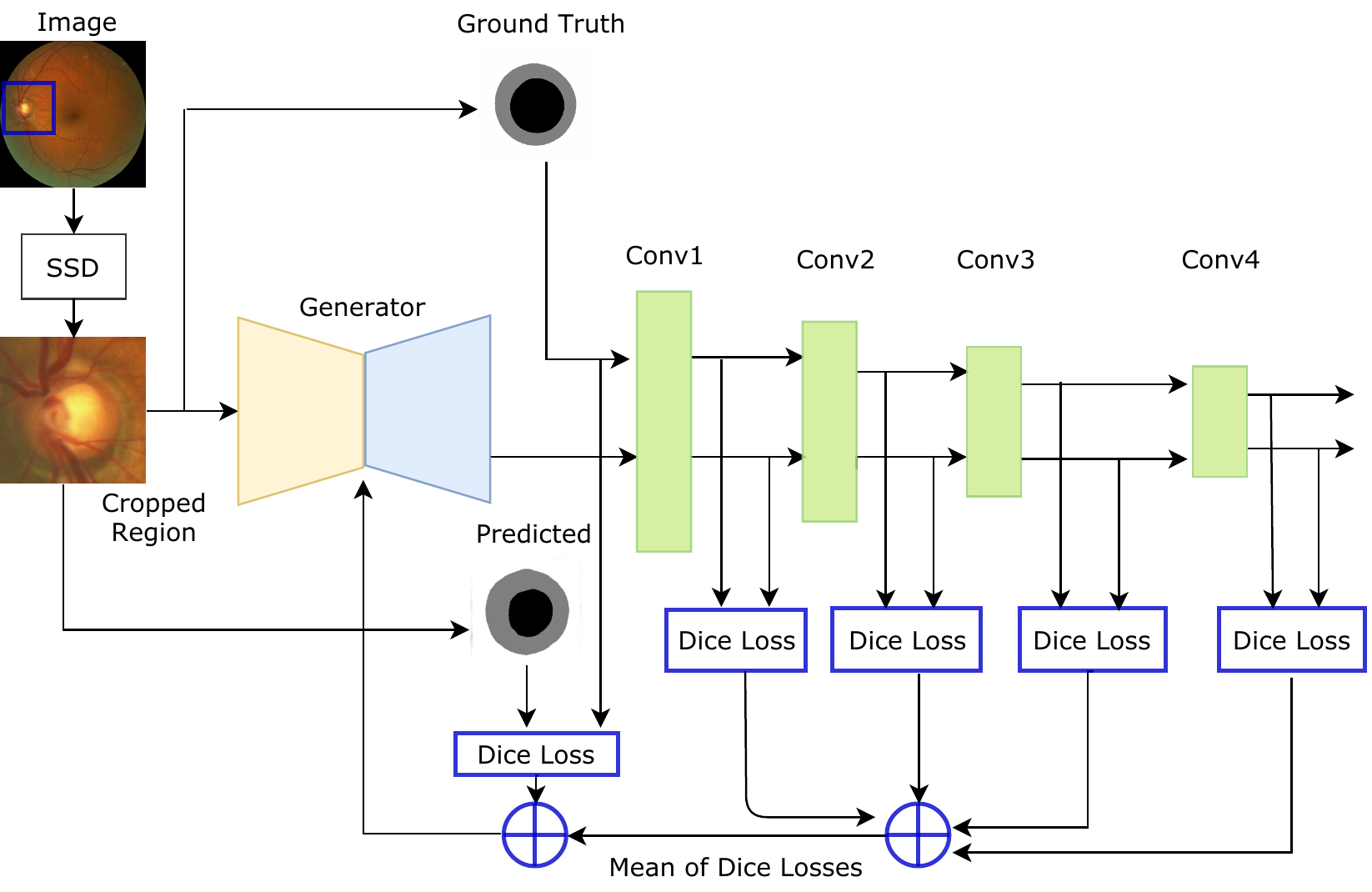}
\caption{Proposed segmentation architecture: generator .}
\label{fig:cGAN architecture}
\end{figure*}

To avoid the overfitting problem, we have applied a data augmentation techniques such as different illumination, scaling, flipping etc to make the data more diverse. 

\subsection{Multi-Scale Feature Matching Segmentation Model }

The ideal CAD system should be able to segment the full image, hence it will automatically locate the OD and OC. However, this is a very difficult task due to the high similarity between pixel distributions of retinal OD and OC. Therefore, removing most of the image non-ROI portions logically helps the model to learn the visual features. The cropped regions provide a balanced proportion between the number of pixels of the three classes (OD, OC and background). Therefore, an approximate frame around the fundus OD and OC regions must be provided and nevertheless the OD and OC segmentation will be very accurate. Thus, our approach is composed of two stages: detection and segmentation.

Firstly, we localize the optic disc and cup regions in an input fundus image using the Single Shot Detector  (SSD) \cite{liu2016ssd}  and video visualization is shown here\footnote{\url{https://youtu.be/miCqw_2eclg}} . The detected Region of Interest (ROI) is then fed to the proposed segmentation model to segment the optic disk and cup areas in the input image. The proposed model is consists of two successive networks. The first network is a generator network that is used for segmenting the input image. This network is an encoder-decoder network with skip connections (U-shape) detailed in in Table \ref{table1}. The second one is a simple CNN network used for extracting the features in multi-scale from the predicted and ground truth images.

The encoder network consists of 8 convolutional layers. The eight layer uses $4\times 4$ convolution to generate 32 feature maps. 
In turn, the 8th layer generates 256 feature maps with a size of $1\times 1$. The weights of the eight layers are randomly initialized. The max pooling with $2\times 2$ with a stride of 2 is used for downsampling the feature maps. In all encoder layers, Leaky-ReLU non-linearities are used with batch normalization to avoid the overfitting and speed-up the training process.

The architecture of the decoder is structured in the same way as the encoder one and includes 8 deconvolutional (e.g., Transpose Convolution) layers, but with a reverse layers ordering, and with downsampling layers being replaced by upsampling layers. The weights of the decoder layers are randomly initialized. All the deconvolutional layers use ReLU functions except the 8th $1\times 1$ deconvolution layer that use Tanh activation to produce the final optic disk and cup segmentation.

The next CNN network is composed of 4 convolutional and downsampling layers. The first layer generates 32 feature maps. Moreover, the 4th layer generates 256 feature maps with a size of $30\times 30$. All convolutions are $4\times 4$ spatial filters applied with a styride of $2$. Their weights are randomly initialized and they use leaky-ReLU functions as activations, in addition to batch normalization.

\begin{table}[htp]
\centering
\caption{Architectural details of the proposed generator netowork}
\label{table1}
\resizebox{\textwidth}{!}{%
\begin{tabular}{|c|c|c|c|c|c|c|c|c|c|}
\hline
Layer Name & Layer Type & K, S, P & Input Size & Output Size & Layer Name & Layer Type & K, S, P & Input Size & Output Size \\ \hline
\multirow{8}{*}{Enoder} & \multirow{8}{*}{\begin{tabular}[c]{@{}c@{}}CONV+\\ BN+\\ Leaky Relu\end{tabular}} & 4, 2, 1 & nx3x256x256 & nx32x128x128 & \multirow{7}{*}{Decoder} & \multirow{7}{*}{\begin{tabular}[c]{@{}c@{}}CT+\\ BN+\\ Relu\end{tabular}} & 4, 2, 1 & nx256x1x1 & nx256x2x2 \\ \cline{3-5} \cline{8-10} 
 &  & 4, 2, 1 & nx32x128x128 & nx64x64x64 &  &  & 4, 2, 1 & nx512x2x2 & nx256x64x64 \\ \cline{3-5} \cline{8-10} 
 &  & 4, 2, 1 & nx64x64x64 & nx128x32x32 &  &  & 4, 2, 1 & nx512x4x4 & nx256x32x32 \\ \cline{3-5} \cline{8-10} 
 &  & 4, 2, 1 & nx128x32x32 & nx256x16x16 &  &  & 4, 2, 1 & nx512x8x8 & nx256x16x16 \\ \cline{3-5} \cline{8-10} 
 &  & 4, 2, 1 & nx256x16x16 & nx256x8x8 &  &  & 4, 2, 1 & nx512x16x16 & nx128x8x8 \\ \cline{3-5} \cline{8-10} 
 &  & 4, 2, 1 & nx256x8x8 & nx256x4x4 &  &  & 4, 2, 1 & nx256x32x32 & nx64x4x4 \\ \cline{3-5} \cline{8-10} 
 &  & 4, 2, 1 & nx256x4x4 & nx256x2x2 &  &  & 4, 2, 1 & nx128x64x64 & nx32x2x2 \\ \cline{3-10} 
 &  & 4, 2, 1 & nx256x1x1 & nx256x1x1 & Output & Tanh & 4, 2, 1 & nx64x128x128 & nx1x256x256 \\ \hline
\end{tabular}%
}
\end{table}

The proposed model has been trained over a loss function resulting from combining a content and multi-scale feature matching losses. The content loss follows a classical approach in which the predicted image is pixel-wise compared with the corresponding one from ground-truth. In turn, the feature matching loss depends of the feature extraction over the ground-truth and the predicted images with observing the input image as a condition for color and texture features. 

Given an input $x$ (a ROI of a fundus image), the U-Net network $G$ represents the generated image $\hat{y}$ as a vector of probabilities of each pixel. The content loss function $\ell_{dice}(G)$ is computed between $\hat{y}$ and its corresponding ground-truth $y$. To maximize the intersection between the two images, the Dice coefficient is used as a content loss in our model that is defined as:

\begin{equation}
    \ell_{dice}(y,G(x))= 1- dice(y, \hat{y}) = 1- \frac{2 |y| . | \hat{y} |}{|y|^2 + | \hat{y} |^2} , \label{equation:dc}
\end{equation}

In turn, the multi-scale feature matching (mfm) loss is also based on the dice loss function, but it is done between the features extracted per layer to compare the features resulted with the ground truth and the predicted image in multi-scale. That error can be defined as:

\begin{equation}
  \ell_{mfm} (x, y, G(x))= \sum_{i=1}^{N} \sum_{c=1}^{H} 1- dice(Y_{ci}, \hat{Y}_{ci}), 
 \label{equation:dc}
\end{equation}
where $H$ is the number of channels per feature maps, $N$ is the number of layer in the CNN network, $Y_{ci}$ is the feature maps with the ground-truth $y$ and $\hat{Y}_{ci}$ is the feature maps with the predicted image $\hat{y}$.

Thus, the final loss function during training is formulated as:

\begin{equation}
    \ell_{T}=\ell_{mfm}(x,y,G(x))+ \lambda \ell_{dice}(y,G(x)), \label{equation:dc}
\end{equation}

This loss $\ell_{T}$ is efficiently integrated into back-propagation of the network through the ADAM optimization.

\section{Experiments and Results}

In the experiments, we used a 64-bit I7-6700, 3.40GHz CPU with 16GB of RAM and NVIDIA GTX 1070 GPU with 8GB of video RAM, running on Ubuntu 16.04 Linux operating system. We used Pytorch\footnote{\url{https://pytorch.org/}} neural network library to devise the proposed neural network model.

We have used 360 images for training and 40 images along with their ground truth has been used to validate our model. During training, we have also tested different learning rates and loss optimizers (SGD, AdaGrad, Adadelta, RMSProp, Adam), finding Adam with $\beta_1$ = 0.5, $\beta_2$ = 0.999 and initial learning rate = 0.0002 with batch size 8 the best combination. For $\lambda$ , we have found experimentally that 150 is the most suited value. Finally, results were achieved by training a generator and CNN as a multi-scale feature extractor from scratch for 10 epochs.

\label{subsec:output}
\begin{figure*}[htp]
\centering
\includegraphics[width=1.0 \textwidth, height=0.4\textheight]{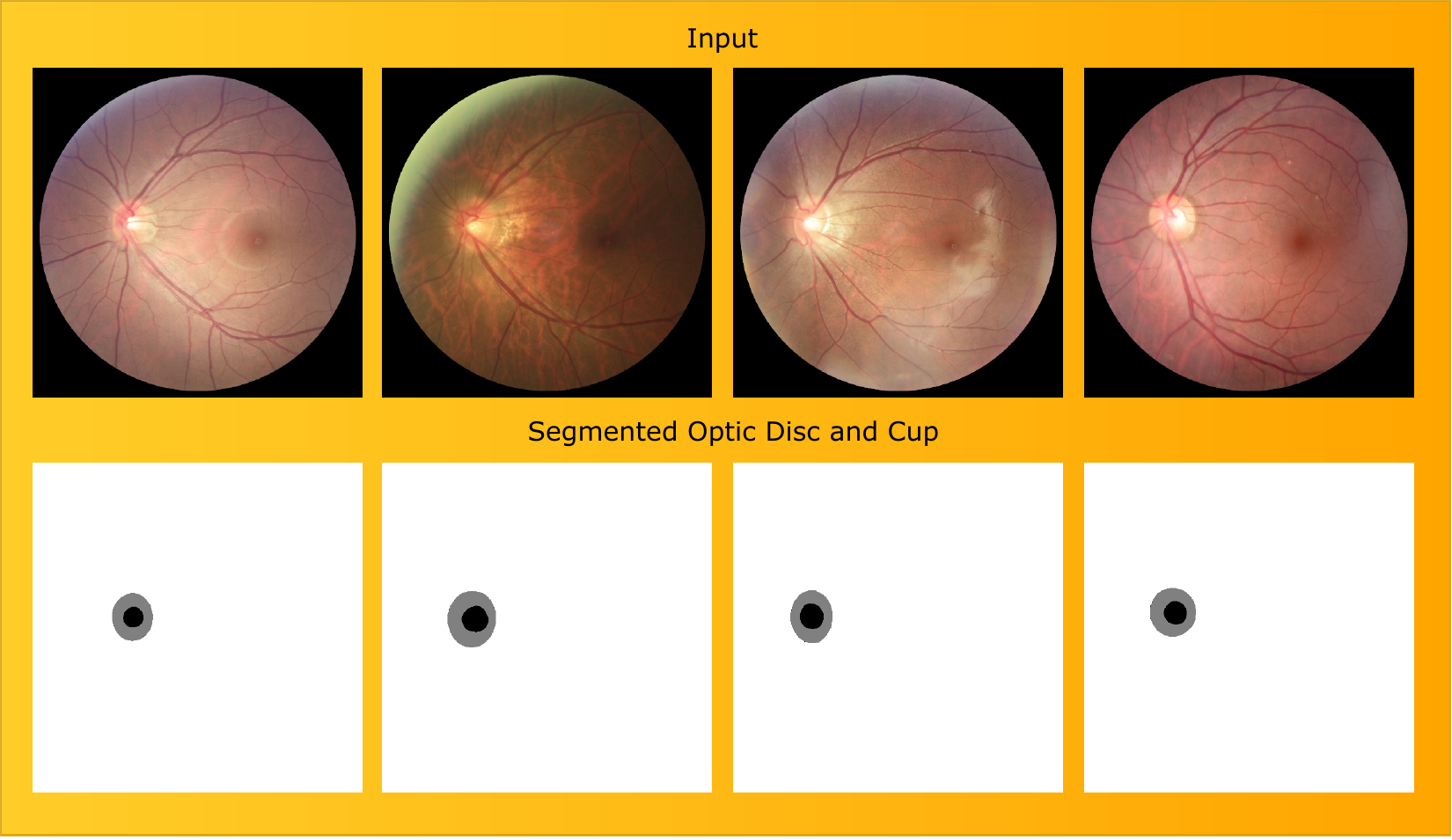}
\caption{Segmentation output of OD and OC from our proposed segmentation model}
\label{fig:cGAN architecture}
\end{figure*}

In Table \ref{table2}, the performance of the proposed model is shown in terms of dice coefficient accuracy for OD and OC segmentation, and MAE of curve to disc ratio (CDR). However, the result shown in Table \ref{table1} is from our second last submission (last submission result was not provided by the organizer, which we believe will outperform our current result). 

The main advantage of the proposed model is, it takes only 10 epoch to train and efficiently segment the OD and OC from the test data.

In Fig \ref{subsec:output} and video visualization \footnote{\url{https://youtu.be/_aIwQphCDeQ}}, shows the segmentation output of OD and OC of a fundus images from the proposed segmentation model. In first row shows an input image and second rows showing a corresponding segmented images (OD in a gray color, OC in a black color and white pixels are for the background).
\begin{table}[htp]
\centering
\caption{Optic Disc and Cup segmentation Dice coefficient accuracy and MAE CDR  with the REFUGE challenge dataset of 400 validation images}
\label{table2}
\scalebox{1.0}{
\begin{tabular}{|c|c|}
\hline
Methods & Accuracy (\%) \\ \hline \hline
       Optic Cup &   ~$ 0.8341\% $         \\ \hline
      Optic Disc  &    ~$ 0.9340\%$          \\ \hline
     MAE CDR &   ~$ 0.0605\% $         \\ \hline
       
\end{tabular}
}
\end{table}

\section{Conclusion}

In this work, we a proposed a U-Net network followed a multi-scale feature matching network to segment the OD and OC in fundus images. A dice loss function is used based on matching the features of the predicted and ground truth images by minimizing the error at each layer. Our proposed model is properly able to distinguish between the two classes of OD and OC regions. The segmentation output of the proposed model can accurately preserve the boundaries of the two regions.

\bibliographystyle{splncs}
\bibliography{biblography} 

\end{document}